\documentclass{article}
\usepackage{ijcai22}

\usepackage{fancyhdr}

\usepackage{times}
\usepackage{soul}
\usepackage{url}
\usepackage[hidelinks]{hyperref}
\usepackage[utf8]{inputenc}
\usepackage[small]{caption}
\usepackage{graphicx}
\usepackage{amsmath,amssymb,amsfonts}
\usepackage{amsthm}
\usepackage{booktabs}
\usepackage{algorithm}
\usepackage{algorithmic}
\usepackage[hindi,main=english]{babel}
\title{Am I No Good? Towards Detecting Perceived Burdensomeness and Thwarted Belongingness from Suicide Notes}

\author{
Soumitra Ghosh$^1$
\and
Asif Ekbal$^1$\and
Pushpak Bhattacharyya$^2$
\affiliations
$^{1}$Department of Computer Science and Engineering, IIT Patna, India\\
$^{2}$Department of Computer Science and Engineering, IIT Bombay, India
\emails
ghosh.soumitra2@gmail.com,
asif@iitp.ac.in,
pb@cse.iitb.ac.in
}

\begin{document}

\maketitle
\thispagestyle{fancy}

\begin{abstract}
The World Health Organization (WHO) has emphasized the importance of significantly accelerating suicide prevention efforts to fulfill the United Nations' Sustainable Development Goal (SDG) objective of 2030. 
In this paper, we present an end-to-end multitask system to address a novel task of detection of two interpersonal risk factors of suicide, \textit{Perceived Burdensomeness (PB)} and \textit{Thwarted Belongingness (TB)} from suicide notes. We also introduce a manually translated code-mixed suicide notes corpus, \textit{CoMCEASE-v2.0}, based on the benchmark CEASE-v2.0 dataset, annotated with temporal orientation, \textit{PB} and \textit{TB} labels. We exploit the temporal orientation and emotion information in the suicide notes to boost overall performance. For comprehensive evaluation of our proposed method, we compare it to several state-of-the-art approaches on the existing CEASE-v2.0 dataset and the newly announced CoMCEASE-v2.0 dataset. Empirical evaluation suggests that temporal and emotional information can substantially improve the detection of \textit{PB} and \textit{TB}.
\end{abstract}

\section{Introduction}
According to the World Health Organization (WHO), more than 700,000 people, or one in every 100, committed suicide in 2019. In 2019, the global suicide rate per 100,000 people was 9.0, with 11.2 in the Africa region, 10.5 in Europe, and 10.2 in Southeast Asia. The Eastern Mediterranean area had the lowest rate (6.4). While global suicide rates fell by 36\% between 2000 and 2019, the Americas region witnessed a 17\% increase\footnote{\url{https://news.un.org/en/story/2021/06/1094212}}. The worldwide catastrophe produced by the COVID-19 pandemic has amplified risk factors for suicidal behaviour, making suicide prevention even more vital now. 

According to the interpersonal theory of suicide \cite{joiner2005people}, suicidal desire arises when individuals experience persistent emotions of \textit{perceived burdensomeness (PB)} and \textit{thwarted belongingness (TB)}. 
The feeling of being a burden on friends, family, and/or society, as well as the potentially dangerous belief that one's death is worth more than one's life, is referred to as \textit{PB}. \textit{TB}, on the other hand, is also a risk factor for developing the impulse to commit suicide if someone feels isolated from friends, family, or other important social networks. Identifying latent vulnerability factors that raise the likelihood of suicide conduct might help determine preventive efforts.

Suicidal behavior is a challenging problem to investigate,
requiring huge samples because of the low baseline rates of suicide attempts and deaths in the general population.
Most importantly, individuals dying by suicide hinder the employment of any methods for psychological assessments by researchers. In such a case, a suicide note might be a valuable asset in attempting to assess an individual's specific personality status and mind rationale \cite{pestian2008classification}.

\begin{itemize}
    \item \textbf{Suicide note excerpt (Hinglish):} \def\DevnagVersion{2.17}{\dn agr m\?rF gStF TF BF to m\?rF} wife \& daughter \def\DevnagVersion{2.17}{\dn ko \7{s}sAiX \3C8wo{\qva} krvAyA gyA{\rs ,\re}}\def\DevnagVersion{2.17} torture {\dn krk\?.}
\item \textbf{English translation:} \textit{Even if it was my fault then why was my wife \& daughter made to commit suicide, by torture.}
\end{itemize}

The usage of code-mixed forms of communication is common in multilingual and multicultural societies worldwide, notably in India, Bangladesh, Pakistan, Singapore, and even the United States of America, the United Kingdom, etc.
The above excerpt is from a suicide note written in Hindi-English code-mixed language. 

Specific time perspective orientations (TPOs), such as having an aversive attitude towards the past (Negative Past) and having a hopeless, nihilistic attitude towards life (Fatalistic Present), are associated with suicide ideation. \cite{shahnaz2019examining} found large and significant differences in TPOs between individuals with a history of suicide ideation and non-suicidal participants. Emotion is also recognised to have an important influence in predicting the author's mental state in a suicide note. The association of various emotions of different polarities with varying time perspectives may help learn the underlying emotional feeling conveyed in the content of suicide notes (SNs). Recent advances in computational approaches may be used to better understand various ideas in psychology research, which might aid in the early detection, evaluation, management, and follow-up of those experiencing suicidal thoughts and behaviors. 

Most existing automated systems need the availability of datasets for training prediction models, and relevant datasets are scarce. Despite the fact that many multilingual speakers in densely populated countries such as India, Pakistan, and Bangladesh use English-Hindi code-mixed language, just a few research \cite{gupta-etal-2018-uncovering,srivastava2020phinc} have attempted to work on this issue. Furthermore, the existing resources \cite{Ghosh2020CEASEAC,ghosh2021multitask} are in English only and annotated at the sentence level, limiting their value in developing automated systems on code-mixed languages and addressing tasks at the document level. During the peak of the COVID-19 epidemic, a large increase in telehealth usage was driven more by persons seeking mental health services than care for physical conditions\footnote{\url{https://www.frontiersin.org/research-topics/25568/artificial-intelligence-in-mental-health}}. Artificial Intelligence (AI)-based automated systems can be incorporated into digital interventions, particularly web and smartphone apps, to enhance user experience and optimise personalised mental health care, more so, in developing countries with high population and minimal healthcare facilities. This motivated us to devise an approach for utilizing the sentence-level information inherent in existing datasets and addressing correlated tasks at the document level by leveraging the underlying connection between them. Also, such an approach can be built on top of current state-of-the-art transformer-based pre-trained models to account for the data scarcity problem and produce good results. 

To this end, the current study presents an automated method to simultaneously detect the presence of two interpersonal states of individuals, \textit{PB} and \textit{TB}, from their SNs, given the temporal orientation information and the emotional states of the note content. We introduce, \textit{CoMCEASE-v2.0}, a manually translated code-mixed suicide notes corpus based on the benchmark CEASE-v2.0 dataset and annotated with temporal orientation, \textit{PB}, and \textit{TB} labels. We compare our proposed technique to various state-of-the-art approaches using the existing CEASE-v2.0 dataset and the recently announced CoMCEASE-v2.0 dataset to provide a complete evaluation. The note-level personality annotations with reference note-ids can be accessed from \url{https://www.iitp.ac.in/~ai-nlp-ml/resources.html#CoMCEASE}.

The main contributions are summarized below:
\begin{itemize}
    \item This is the first study towards detection of \textit{perceived burdensomeness} and \textit{thwarted belongingness} from suicide notes using computational methods.
    \item We introduce a good quality manually translated code-mixed corpus of SNs based on the benchmark CEASE-v2.0 English dataset with manual annotations for \textit{TPO}, \textit{PB} and \textit{TB}. 
    \item We propose a Temporal and Emotion-assisted Multitask Framework (\textit{TEMF}) to jointly detect \textit{PB} and \textit{TB} from SNs.
\end{itemize}


\section{Related Work}\label{relwork}
Several studies have performed a content analysis of SNs and identified distinguishing features that differentiate SNs from other types of texts
\cite{handelman2007content,pestian2012sentiment}. 
Suicidal thoughts has been linked to the concepts of \textit{PB} and \textit{TB} in various research. The frequency of occurrence of \textit{PB} and \textit{TB} in SNs was investigated in \cite{gunn2012thwarted}. 
Gender differences in the determinants and correlates of \textit{PB}, \textit{TB}, and acquired suicidal capacity were investigated in \cite{donker2014gender}. 

Temporal perspective was found to have a significant effect in predicting suicidal thoughts in teenagers \cite{laghi2009suicidal}. 
Notable changes were observed in the time perspective of depressed patients when compared to non-depressed people \cite{lefevre2019time}. 
Analyzing the emotional balance in SNs, \cite{teixeira2021revealing} discovered links between the concepts and emotional states of persons who committed suicide by recreating the knowledge structure of such notes. 


The bulk of suicide note research have concentrated on identifying suicide notes \cite{handelman2007content}, with a few focusing on related risk factors for suicide \cite{ghosh2021multitask}. Furthermore, massive pre-trained transformer-based language models have taken NLP by storm in recent years, delivering state-of-the-art performance on many downstream tasks.
To this end, our study introduces a manually translated code-mixed suicide notes corpus with reliable annotations for \textit{TPO}, \textit{PB}, and \textit{TB}, and use computational techniques to detect two crucial interpersonal risk factors, \textit{PB} and \textit{TB}, using an effective multitask network. 

\section{Dataset}\label{datalabel}
We consider the benchmark CEASE-v2.0 dataset \cite{ghosh2021multitask} for this study. This is the only available emotion annotated corpus of suicide notes publicly available for research purposes to the best of our knowledge. The dataset consists of 4932 sentences from over 350 suicide notes annotated with 15 fine-grained emotions at the sentence level. We used the sentence-to-note reference identifiers (IDs)\footnote{provided by \cite{ghosh2021multitask} on request to use it for research purpose solely} to establish the order of sentences in the dataset, as the original resource was released at sentence-level, all shuffled to prevent reconstruction and preserve anonymity. We perform three main tasks on this existing dataset: (A). Manually annotate for \textit{PB} and \textit{TB} (at the note-level), (B). Manually annotate for \textit{TPO} (at the sentence-level), and, (C). Produce Hindi-English (Hi-En) code-mixed translations for the CEASE-v2.0 English dataset (at the sentence level).

\subsection{Annotator Details}
Each task was performed by three annotators (A1, A2, and A3), two doctoral degree holders, and one graduate student, sufficiently acquainted with labeling tasks and well-versed with the concepts of \textit{PB}, \textit{TB} and \textit{TPO}. The annotators are familiar with the nuances of Hindi-English code-mixed communication. All of them are native Hindi speakers, well-proficient in English, and also use Hi-En code-mixed language in daily communications.

\subsection{Annotations for \textit{PB} and \textit{TB}}
Each suicide note is labeled \textit{PB/Not-PB} and \textit{TB/Not-TB} to indicate the existence of \textit{PB} and \textit{TB}. The final annotations were acquired by using a majority vote on the labels provided by the three annotators. We obtain a Fleiss-Kappa \cite{spitzer1967quantification} agreement of 0.791 and 0.76 for the \textit{PB} and \textit{TB} tasks respectively, which indicates substantial agreement among the annotators. 


\subsection{Annotation for TPO}
We perform the annotations for the temporal orientation at the sentence level as a single note often contains references to various time perspectives. 
Consider the suicide note excerpt below:
\begin{center}
\textit{The rest of my life would only be a burden for others. I am unable to do anything because of poor health.}
\end{center}
We can observe that the first sentence relates to a \textit{future} event while the following sentence relates to a \textit{present} event. 
We follow the connotations of \textit{past}, \textit{present} or \textit{future} as described by \cite{Kamila2018FineGrainedTO}. Any event(s) that has started and ended is labelled as \textit{past} whereas an ongoing event is marked as \textit{present} and event(s) that are yet to happen are labelled as \textit{future}. We obtain a Fleiss-Kappa agreement of 0.745 which indicates that the annotations are of substantial quality.  

\subsection{Code-Mixed Manual Translation of Benchmark CEASE-v2.0 Dataset}
We construct code-mixed (Hindi-English) translations of each English sentence in the CEASE-v2.0 dataset while preserving \textit{fluency (F)} (syntax) and \textit{adequacy (A)} (semantics). The resulting dataset is known as \textit{CoMCEASE-v2.0}. We use the matrix language frame (MLF) theory \cite{joshi1982processing} to create the translations, which states that a code-mixed text has a dominant language or matrix language (here, Hindi) and an inserted language or embedded language (here, English). The insertions could be words or bigger elements that adhere to the matrix language's grammatical structure.
Based on previous works \cite{gupta-etal-2018-uncovering} on code-mixed text creation, we identified three key scopes where the words or components from the original English sentence can be maintained in the translated code-mixed text. These are: Named Entities (NEs) of types 'Person', 'Location' and ‘Organization', noun phrases, and adjective words. However, we need not consider the NEs as the original CEASE-v2.0 dataset comes with anonymized forms of all NEs (as also mentioned in \cite{ghosh2021multitask}). 

Each annotator (say A1) was asked to mark every translated sentence of the next annotator (A2) for both \textit{F} and \textit{A} with an ordinal value from a scale of 1-5\footnote{\textbf{Fluency} - 5: \textit{Flawless}, 4: \textit{Good}, 3: \textit{Non-native}, 2: \textit{Disfluent}, 1: \textit{Incomprehensible}; \textbf{Adequacy} - 5: \textit{All}, 4: \textit{Most}, 3: \textit{Much}, 2: \textit{Little}, 1: \textit{None}}. We obtain high average \textit{F} and \textit{A} scores of 4.31 and 4.65, respectively, which indicates that the translations are of good quality. A couple of sample translations from the \textit{CoMCEASE-v2.0} dataset are shown below: 
\begin{enumerate}
    \item \textbf{Original (En):} \textit{Do not feel sorry for me.}\\
    \textbf{Translation (Hi-En):} \def\DevnagVersion{2.17}{\dn m\?r\? Ele} \def\DevnagVersion{2.17} sorry feel {\dn mt kro.}
    \item \textbf{Original (En):} \textit{I was immediately taken in a wheelchair to the psychiatrist's office.}\\
    \textbf{Translation (Hi-En):} \def\DevnagVersion{2.17}{\dn \7{m}J\? \7{t}r\2t} wheelchair \def\DevnagVersion{2.17}{\dn s\?} psychiatrist \def\DevnagVersion{2.17}{\dn k\?} office \def\DevnagVersion{2.17}{\dn l\? jAyA gyA.}
\end{enumerate}

\begin{table}[ht]
\renewcommand{\arraystretch}{1}
\centering
\scalebox{1}{
\begin{tabular}{c|c|c|c|c}
\hline
\textbf{HE} & \textbf{A1} & \textbf{A2} & \textbf{A3} & \textbf{Average}\\
\hline
\textit{F} & 4.25 & 4.28 & 4.40 & 4.31\\
\textit{A} & 4.45 & 4.80 & 4.71 & 4.65\\
\hline
\end{tabular}
}
\caption{Translation quality scores from human evaluation}\label{transagree}
\end{table}

\begin{figure*}
\centering
\includegraphics[width=17cm]{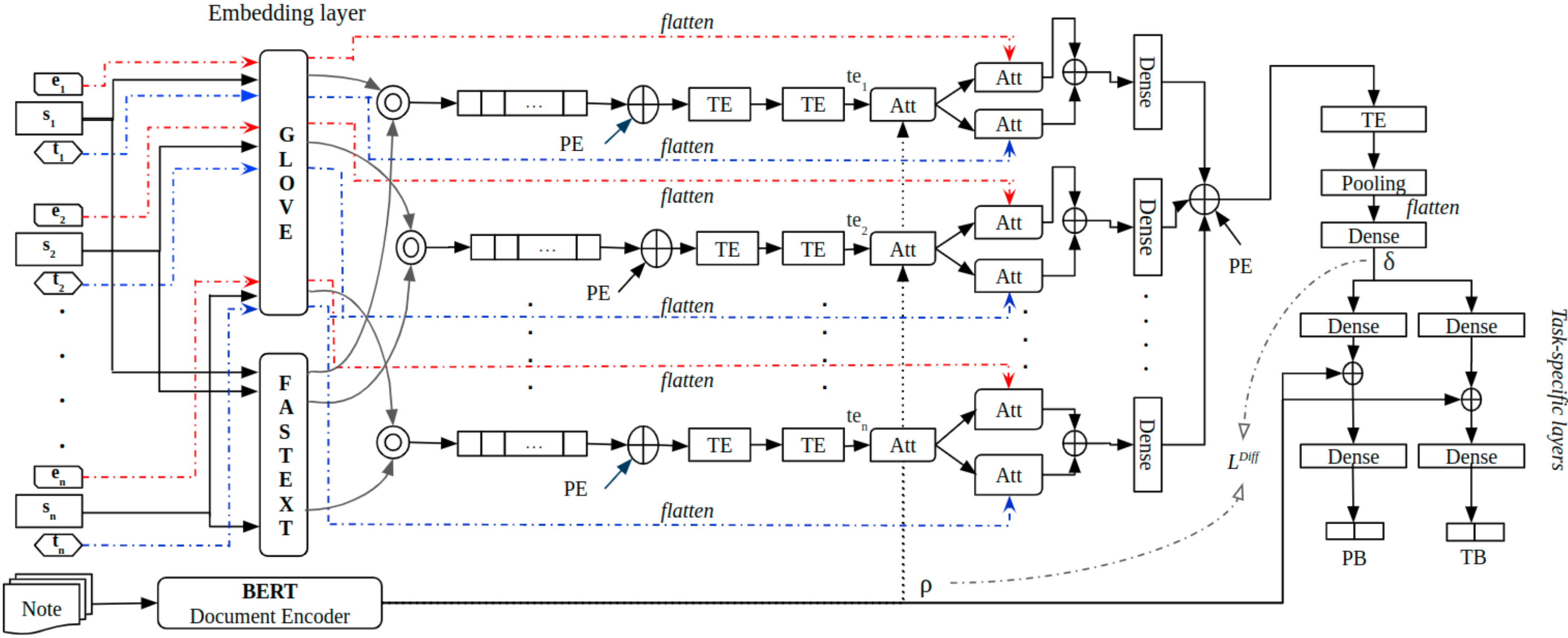}
\caption{Architecture of the Temporal and Emotion-assisted Multitask Framework (TEMF). PE: positional encodings; TE: transformer encoder; Att: additive attention.} \label{proposed}
\end{figure*}

\subsection{Dataset statistics}
Out of the 364 suicide notes in CEASE-v2.0 dataset which we annotated for \textit{PB} and \textit{TB}, 27\% of the notes are marked as either \textit{PB} (10.44\%), \textit{TB} (11.81\%) or both (4.67\%). Total of 55 notes are labelled as \textit{PB} and 60 notes as \textit{TB}. We consider 4885 sentences from the 4932 phrases in the CEASE-v2.0 dataset for the temporality annotations and our experiments. The remaining 47 sentences were not considered due to the following reasons: non-matching mapping IDs (as provided by the authors in \cite{ghosh2021multitask}), confusion among all annotators (while annotating for TPO), single-word sentences. The dataset is distributed over the various temporal categories as follows: \textit{past} - 1508 sentences (30.87\%), \textit{present} - 2035 sentences (41.66\%) and \textit{future} - 1342 sentences (27.47\%). The average sentence length in the \textit{CoMCEASE-v2.0} dataset is 16.73, whereas it is 14.96 in the original dataset.

\section{Methodology}\label{methodlabel}
We present a detailed discussion on the proposed \textit{TEMF} system in the following sub-sections. We depict the overall architecture of the approach in Figure~\ref{proposed}.

\subsection{Task Definition}
Given a suicide note ($S^i$) with sentences annotated with temporal orientation and emotion labels, the task objective is to detect the presence or absence (1 or 0) of \textit{PB} and \textit{TB}. Let $D$ denotes the dataset consisting of the $N$ suicide notes, then $D=(S^1,S^2,...,S^N)$. A note $S^i$ is a collection of sentences such that $S^i = (s_1^i, s_2^i,..., s_n^i)$, where $n$ denotes the number of sentences in the $i^{th}$ note. The emotion labels ($E^i$) and temporality labels ($T^i$) corresponding to each sentence in $S^i$ is denoted as $E^i = (e_1^i, e_2^i,..., e_n^i)$ and $T^i = (t_1^i, t_2^i,..., t_n^i)$, respectively. The proposed approach work towards maximising the value of the following function:
\begin{multline}
 \underset{\theta}{argmax} (\Pi_{i=0}^N P(y_{PB}^i,y_{TB}^i | S^i, E^i, T^i; \theta))
\end{multline}
where $y$ $\epsilon$ $\{1,0\}$ and $\theta$ denotes the model parameters to be optimized.

\subsection{Temporal and Emotion-assisted Multi-task Framework (TEMF)}
This section discusses our proposed system, which comprises the six main components: a). Input Embeddings, b). Document Encoder, c). Sentence Encoder, d). Attention Module e). High-level Document Abstraction, f). Task-specific Dense layers and Output Layers. 

\textbf{Input Embeddings: }
Since our focus is on working with the Hinglish dataset, we consider the pre-trained GloVe\footnote{\url{https://nlp.stanford.edu/data/glove.840B.300d.zip}} (English) and fastText\footnote{\url{https://fasttext.cc/docs/en/crawl-vectors.html}} (Hindi) word embeddings to capture the semantics of the words of the input sentences. 
We embed the emotion class and temporality class labels with the word vectors ($\mathbb{R}^{1 \times D}$) from GloVe embeddings as all the class labels are in English. \textit{D} is the dimension of the embedding (in our case, 300). 

The sentence embedding representations from GloVe and fastText are projected to a single vector space by finding their mathematical mean, as inspired by the work in \cite{Wang2019YNUWBAS}. Each $s^i$ is embedded as $W_{emb}^i$ = $\{w_1, w_2, ..., w_c\}$, \textit{W} $\epsilon$ \textit{c $\times$ d}, where \textit{c} is the length of $s^i$. 

\textbf{Document Encoder:} We use the BERT (\textit{base}) \cite{devlin-etal-2019-bert} pre-trained model to encode the information of a suicide note at the document level. The representation from the $[CLS]$ token of the BERT encoder's last encoder layer encodes the whole document's contextual information. 

\textbf{Sentence Encoder:} At the sentence level, we employ the transformer encoder (TE) \cite{vaswani2017attention} to capture the local semantic information as well as long-term dependencies. First, the positional encodings ($PE^i$) and the word embeddings ($W_{emb}^i$) are added to generate the input embeddings ($IE^i = W_{emb}^i + PE^i$). This allows the model to use word-order information, including relative and absolute positional data. Each transformer module performs a multi-head self-attention (MHA) operation on the input, followed by a fully connected point-wise feed forward network (FFN) to produce knowledge representation, passing the embedding representation of each sentence $IE^i$ through two successive layers of transformer encoder. 
\begin{gather}
\footnotesize R^{(l)^i} = \text{MHA}(IE^i, IE^i, IE^i)\\
\footnotesize S^{(l)^i} = \text{FFN}(R^{(l)^i})
\end{gather}
where $l$ is the number of layers in the network.

\textbf{Attention Module: }
We infuse the document-level contextual information from BERT ($\rho$) to the sentence-level transformer outputs ($te_i$) by performing additive-attention (Att) \cite{bahdanau2015neural} to make each sentence aware of the global context. The attention-mechanism can be realized by the following equations: 
\begin{gather}
\footnotesize \gamma = W_3^T\text{tanh}(W_1{\rho} + W_2{te_i^c})\\
\footnotesize {{\alpha}_i = \frac{exp(\gamma({{\rho}{te_i^c}}))}{\sum_{j=1}^{c}exp(\gamma({\rho}{te_j^c}))}}
    \\
\footnotesize {\phi_{i} = \sum_{t=1}^{c} {{\alpha}_i}{te_i^c}}
\end{gather}

where $W_1$, $W_2$, $W_3$ are the learnable weight matrices, $tanh$ is a non-linear function.

We further apply two independent additive attention operations on each context-aware sentence output, one with the temporal embedding as the query vector and the other with the emotion class embedding as the query vector. 
These attention mechanisms enable to focus on the relevant words in the sentence concerning each sentence's temporal and emotional information. The temporal and emotion-aware representations of each sentence are concatenated and passed through a sentence-specific dense layer.

\textbf{High-level Document Abstraction: }
The dense representations give an abstract representation of the input sentences, which are concatenated ($\oplus$) together and passed through another TE to learn the temporal and emotion-aware global context information. The sentence-level position embedding is also added with the input to the TE to use the sentence-order information. Hierarchical transformer encoders have shown promising output while capturing document-level contextual information. We apply a max-pooling operation on the transformer output ($\delta$) to capture the essential latent semantic information across the input. 

\textbf{Task-specific Dense layers and Output Layers: }
The max-pooled output ($\Delta$) is passed to two task-specific dense layers that capture the intermediate features particular to each task, followed by two task-specific fully connected layers with \textit{softmax} activation function that acts as the output classification layers. The BERT document representation is added to the task-specific intermediate representations to enhance each feature set.

\emph{Calculation of loss:} We train the model through a unified loss function as shown below:
\begin{equation}
 \footnotesize    { L = \alpha * L^{PB} + \beta * L^{TB}} + L^{Diff}
\end{equation}
Where $L^{PB}$ and $L^{TB}$ are the categorical cross-entropy losses, and $\alpha$ and $\beta$ are the loss weights for the two tasks. 

Alongside the cross-entropy losses (\textit{L\textsuperscript{T}}), we calculate a differential loss to minimize the mean squared difference ($L^{Diff}$) between the output representations of BERT and TE to build an adequate shared feature space. The intuition is to reduce the vulnerability mainly due to two reasons: (1) the variation in different encoding methods and (2) handling two document representations (one generated from the whole document and one generated from constituent sentences. 
\begin{equation}
\footnotesize     L^T = -\frac{1}{N} \sum_{i=1}^{N} t_i log O^T \ \ \mathrm{and} \ \ \mathrm L^{Diff} = \sum_{b=1}^{n}(\delta_i-\rho_i)^2
\end{equation}
where $t_i$ $\epsilon$ $\{0, 1\}$ is the ground truth label of each task and \textit{T} denotes the two tasks. 

\section{Experiments and Results}\label{reslabel}
In this section, we discuss the experiments performed and the results and analysis.

\subsection{Baselines}
We evaluate the efficacy of our proposed method on the CEASE-v2.0 dataset considering the following state-of-the-art systems as baselines: Convolutional Neural Network \cite{kim-2014-convolutional}, Convolutional Neural Network+Context Long Short Term Memory (CNN+cLSTM) \cite{poria2017context}, BERT \cite{devlin-etal-2019-bert} and CMSEKI \cite{ghosh2021multitask}. In CNN+cLSTM system, CNN is used for feature extraction at utterance level followed by a cLSTM to learn context-aware utterance representations. The CMSEKI system was introduced in the work that presented CEASE-v2.0 dataset, addressing detection of depression, sentiment and emotion, using common-sense knowledge. We adapted the CMSEKI system to address our \textit{PB} and \textit{TB} tasks. For evaluation on the \textit{CoMCEASE-v2.0} dataset, we consider the following state-of-the-art systems as baselines: (1) K-Max Pooling CNN (k-max CNN) \cite{Wang2019YNUWBAS} that uses meta-embeddings formed from GloVe and fastText, (2) multilingual BERT (mBERT) \cite{devlin-etal-2019-bert} pre-trained model.

\subsection{Implementation Details} 
We used 10-fold cross-validation on both datasets and the macro-F1 measure to evaluate all models, as the data is highly skewed over the various classes. For our experiments, we chose the average note length of 13 as the document length. The sequence lengths were set to the average sentence lengths of 15 and 17 for the CEASE-v2.0 and \textit{CoMCEASE-v2.0} datasets, respectively. For each TE, we used 5 self-attention heads with an embedding dimension of 300 and a feed-forward dimension of 600. All models were trained using a batch size of 4 (to maximise GPU use) and 6 epochs. We chose a learning rate of 2e-5 for the \textit{TEMF\textsuperscript{MT}} system. The Grid Search approach was used to fine-tune parameters. The best model on the validation set for each fold were saved for testing. We used the BERT\textsubscript{Base} (English and multilingual) pre-trained models\footnote{\url{https://github.com/google-research/bert}}. All the dense layers (except the output dense, which uses softmax activation) used ReLU activation. The task-specific fully-connected and output layers have 128 and 2 neurons each for each task. We trained all the models with Adam optimizer using backpropagation. The implementations were done on an NVIDIA GeForce RTX 2080 Ti GPU. To account for the non-determinism of different TensorFlow GPU operations, we report F1 scores averaged across the five 10-fold cross-validation runs.

\subsection{Results and Analysis}
We observe from Table~\ref{res} that the proposed \textit{TEMF\textsuperscript{MT}} system outperforms all the baselines commendably with improvements of 2.86\% (on \textit{PB} Task) and 3.08\% (on \textit{TB} task) from the next best performing models BERT\textsubscript{Base}\textsuperscript{ST} and CMSEKI\textsuperscript{MT}, respectively, on the CEASE-v2.0 dataset. The \textit{TEMF\textsuperscript{MT}} system also obtains highest F1-scores of 51.27\% and 55.21\% on the CoM-CEASE-v2.0 dataset, with improvements of 2.34\% (on \textit{PB} Task) and 4.12\% (on \textit{TB} task) from the next best performing model mBERT\textsubscript{Base}\textsuperscript{ST}. The \textit{PB} task yielded higher scores for all models than the \textit{TB} task, which might be due to a slightly higher number of \textit{TB} occurrences in the dataset than \textit{PB}.

\begin{table}
\renewcommand{\arraystretch}{1.4}
        \centering
\scalebox{0.98}{
        \begin{tabular}{c|c|c}
        \hline
         Models &  F1\textsuperscript{PB} (\%) &  F1\textsuperscript{TB} (\%) \\
        \hline
         CNN\textsuperscript{ST} \cite{kim-2014-convolutional} &  46.6 &  50.4\\
         CNN+cLSTM\textsuperscript{ST} \cite{poria2017context} &  45.4 &  45.8\\
         BERT\textsubscript{Base}\textsuperscript{ST} \cite{devlin-etal-2019-bert} &  51.16 &  52.02\\
         CMSEKI\textsuperscript{MT} \cite{ghosh2021multitask} &  49.68 &  53.56\\
        \hline
         TEMF\textsuperscript{MT} \textit{(proposed)} &  \textbf{54.02} &  \textbf{56.64}\\
        \hline
         TMF\textsuperscript{MT} \textit{(ablation 1)} &  53.33 &  52.43\\
         EMF\textsuperscript{MT} \textit{(ablation 2)} &  51.63 &  48\\
        \hline
       \end{tabular}}
       \caption{Results on the CEASE-v2.0 dataset. Here, ST and MT denote single-task and multi-task system, respectively.}
       \label{res}
    \end{table}
    
\begin{table}
\renewcommand{\arraystretch}{1.4}
        \centering
\scalebox{0.98}{
        \begin{tabular}{c|c|c}
        \hline
         Models &  F1\textsuperscript{PB} (\%) &  F1\textsuperscript{TB} (\%) \\
        \hline
         k-max CNN \cite{Wang2019YNUWBAS} &  47.43 &  51.06\\
         mBERT\textsubscript{Base}\textsuperscript{ST} \cite{devlin-etal-2019-bert} &  48.93 &  51.09\\
        \hline
         TEMF\textsuperscript{MT} \textit{(proposed)} &  \textbf{51.27} &  \textbf{55.21}\\
        \hline
         TMF\textsuperscript{MT} \textit{(ablation 1)} &  50.03 &  47.01\\
         EMF\textsuperscript{MT} \textit{(ablation 2)} &  49.37 &  47.02\\
        \hline
       \end{tabular}}
       \caption{Results on \textit{CoMCEASE-v2.0} dataset}
       \label{comres}
    \end{table}

\begin{figure}[ht]
\centering
\includegraphics[scale=0.35]{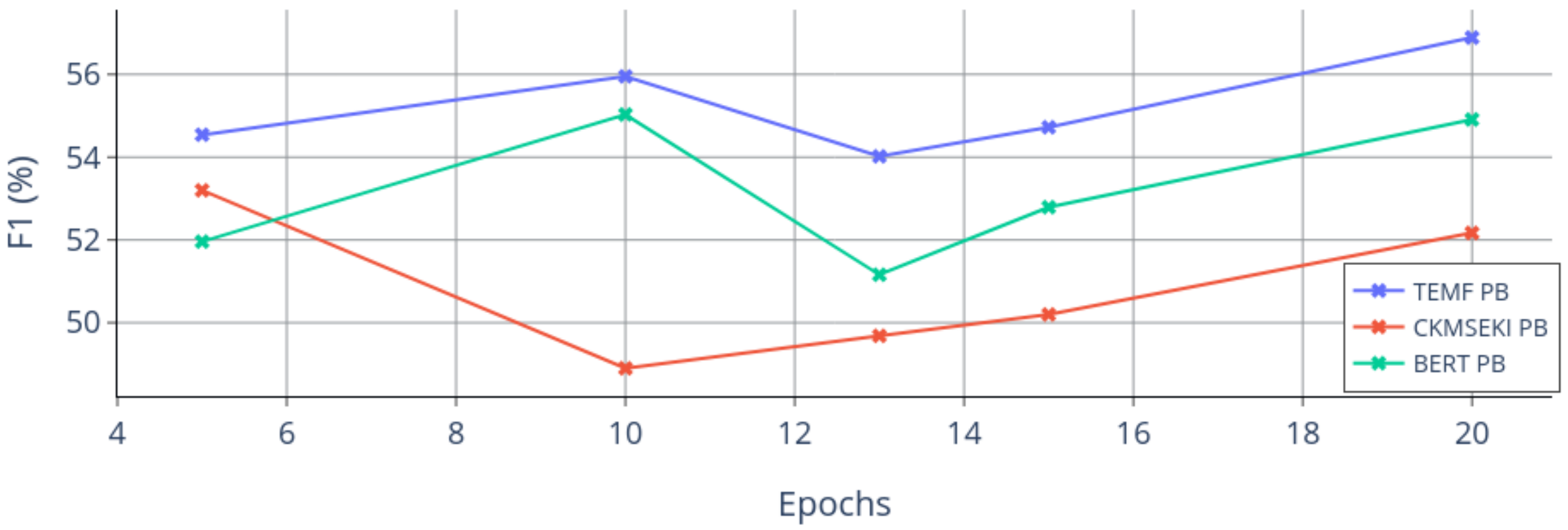}
\caption{Graphical depiction of results of \textit{TEMF} on varying context length of suicide notes.} \label{cxtlenpb}
\end{figure}

\begin{figure}[ht]
\centering
\includegraphics[scale=0.35]{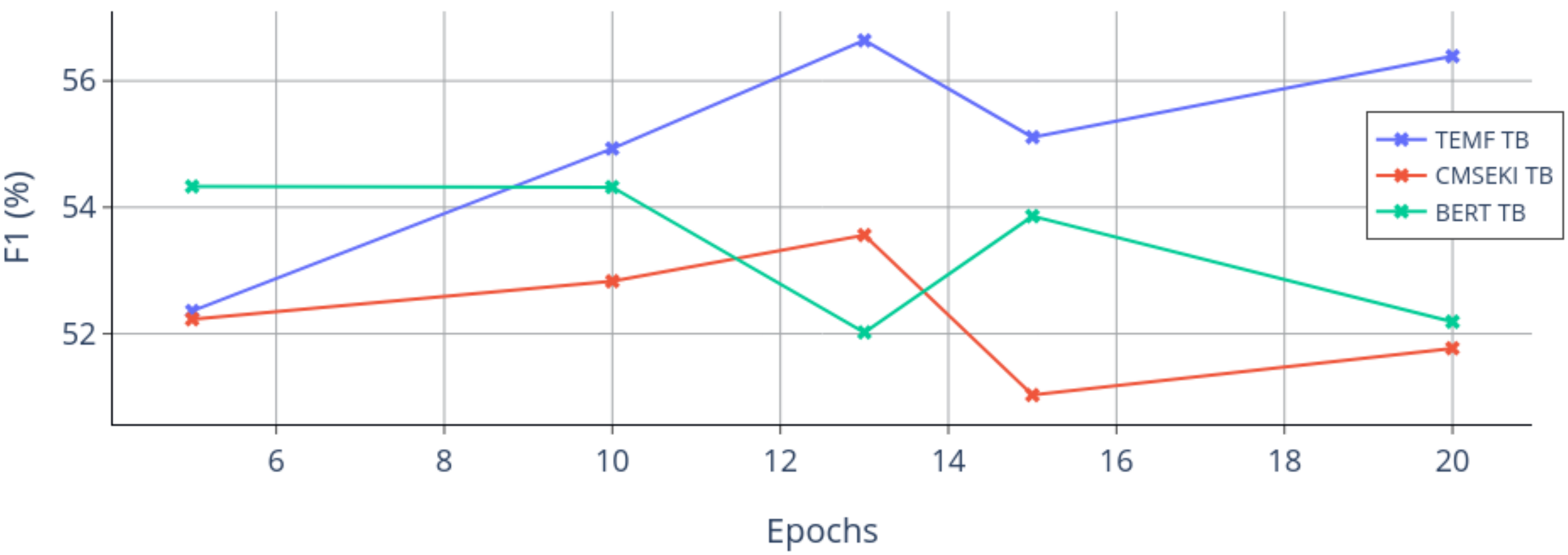}
\caption{Graphical depiction of results of \textit{TEMF} on varying context length of suicide notes.} \label{cxtlentb}
\end{figure}

\textbf{Varying Context Length: } 
Figure~\ref{cxtlenpb} and ~\ref{cxtlentb} depicts the F1 results of our \textit{TEMF} system with comparison to the best performing baselines on different context lengths (5, 10, 13, 15, 20). We observe an upward trend towards the improvement of \textit{PB} and \textit{TB} scores by our \textit{TEMF\textsuperscript{MT}} system, outperforming the CMSEKI\textsuperscript{MT} and BERT\textsuperscript{MT} systems, as contextual information is increased. Furthermore, on the TB task, unlike the BERT\textsuperscript{MT} and CMSEKI\textsuperscript{MT} systems where performance deteriorates with the increase in context length, the proposed \textit{TEMF\textsuperscript{MT}} model improves substantially.


\textbf{Comparison with Prior Works: }
We observe from Tables~\ref{res} and ~\ref{comres} that our proposed \textit{TEMF\textsuperscript{MT}} system outperforms the considered state-of-the-art baselines on both the CEASE-v2.0 and CoM-CEASE-v.0 datasets. The best performing baselines on the \textit{PB} and \textit{TB} tasks are BERT\textsubscript{Base}\textsuperscript{ST} and CMSEKI\textsuperscript{MT}, respectively for the CEASE-v2.0 dataset and mBERT\textsubscript{Base}\textsuperscript{ST} for the CoMCEASE-v2.0 dataset. Moreover, the poor performance of the various baselines on both datasets show how critical it is for present context modelling approaches to perceive interpersonal content from texts. 

\textbf{Ablation Study: }
We modify TEMF\textsuperscript{MT} by removing the temporal orientation labels and emotion labels in the input, one at a time, to develop EMF\textsuperscript{MT} and TMF\textsuperscript{MT}, respectively. Tables~\ref{res} and \ref{comres} show that eliminating either one of the time or emotional information from the input has a significant detrimental influence on the overall performance of the model for both the datasets. Furthermore, the results show that removing the temporal information from the input has a more significant impact than removing the emotion data.

\textbf{Error Analysis: }
We identify some circumstances in which our proposed technique fails to accurately categorize \textit{PB} and \textit{TB} instances.
\begin{enumerate}
    \item \textbf{Inconsistent grammar rules:} Writing style in suicide notes is often informal and does not follow the standard rules of sentence structure. This makes it difficult for our model to recognize syntactic information correctly. For example, \textit{Kept me in a locked shower stall for 1 hour. Next sent me burnt food. Giant bugs crawling over my hands. NO FUN!!}
    \item \textbf{Insufficient context:} Lack of knowledge of the context
makes the task of detection for \textit{PB} and \textit{TB} challenging. For example, \textit{They tried to get me, I got them first!}
    \item \textbf{Sarcastic undertone:} The proposed method finds it hard to capture sarcasm, and the problem becomes even critical with lack of sufficient context. For example, \textit{It's amazing to look at my grade sheet.}
\end{enumerate}


\section{Conclusion}\label{conlabel}

In this work, we proposed a Temporal and Emotion-aware Multitask Framework (TEMF) for joint detection of two interpersonal risk factors of suicide: \textit{perceived burdensomeness} and \textit{thwarted belongingness}, from suicide notes. We introduced a manually translated code-mixed Hindi-English corpus (\textit{CoMCEASE-v2.0}) of suicide notes with manual annotations for \textit{TPO}, \textit{PB} and \textit{TB}. 
We empirically observe that the infusion of temporal and emotion information has a positive impact on improving the performances of the \textit{PB} and \textit{TB} detection tasks from suicide notes. Also, the proposed method enables utilizing the existing sentence-level CEASE-v2.0 dataset to address note-level tasks.
 
Future study into the potential pathways that increase the ability to commit suicide would be beneficial as well, such as duration, intensity, and frequency for \textit{PB}, \textit{TB} and acquired capability for self-injury.

\section*{Ethical Consideration}
Our resource creation utilizes publicly available CEASE-v2.0 \cite{ghosh2021multitask} benchmark suicide notes dataset. We followed the data usage restrictions and did not violate any copyright issues. This study was also evaluated and approved by our Institutional Review Board (IRB). We shall make the code and data available for research purposes (on acceptance), through appropriate data agreement procedure.

\section*{Acknowledgement}
Asif Ekbal acknowledges the Young Faculty Research Fellowship (YFRF), supported by Visvesvaraya PhD scheme for Electronics and IT, Ministry of Electronics and Information Technology (MeitY), Government of India, being implemented by Digital India Corporation (formerly Media Lab Asia).

\bibliography{ijcai22}
\bibliographystyle{named}

\end{document}